# Lattice based Conceptual Spaces to Explore Cognitive Functionalities for Prosthetic Arm

M.S. Ishwarya        Ch. Aswani Kumar
School of Information Technology and Engineering
Vellore Institute of Technology & Engineering, Vellore, India.
Email: cherukuri@acm.org
(dated: 11.4.2018)

**Conflict of Interest**

Aswani Kumar Cherukuri and M S Ishwarya have received research grant from Department of Science & Technology, Govt. of India under the Cognitive Science Research Initiative (CSRI) SR/CSRI/ 118/2014. Aswani Kumar Cherukuri declares that he has no conflict of interest. M S Ishwarya declares that she has no conflict of interest

# Lattice based Conceptual Spaces to Explore Cognitive Functionalities for Prosthetic Arm

**Abstract**

**Introduction**- Upper limb Prosthetic can be viewed as an independent cognitive system in order to develop a conceptual space. In this paper, we provide a detailed analogical reasoning of prosthetic arm to build the conceptual spaces with the help of the theory called geometric framework of conceptual spaces proposed by Gärdenfors.

**Method**- Terminologies of conceptual spaces such as concepts, similarities, properties, quality dimensions and prototype are applied for a specific prosthetic system and conceptual space is built for prosthetic arm. Concept lattice traversals are used on the lattice represented conceptual spaces.

**Results**- Cognitive functionalities such as generalization (Similarities) and specialization (Differences) are achieved in the lattice represented conceptual space.

**Conclusions**- This might well prove to design intelligent prosthetics to assist challenged humans. Geometric framework of conceptual spaces holds similar concepts closer in geometric structures in a way similar to concept lattices. Hence, we also propose to use concept lattice to represent concepts of geometric framework of conceptual spaces. Also, we extend our discussion with our insights on conceptual spaces of bidirectional hand prosthetics.

**Keywords:** Action representation, Cognitive system, Conceptual spaces, Concepts, Grasp, Lattice, Prosthetics.

## 1. Introduction:

Cognitive systems are characterized by receiving information and cues from external environment, sensory receptors, responding for cues, Learning, memorizing, categorizing information [1]. As for now, several analogies were derived for different neuro- psychological systems to cognitive system [2]. Treating a neuro-psychological system as a cognitive system would lead to better understanding of neuro- psychological system and might possibly lead to better aid to assist the defects in the system [2]. In this paper, we extend this attempt to a prosthetic system. We treat the prosthetic system as a cognitive system with a view for better understanding and improvement by developing conceptual spaces. In order to develop conceptual spaces of prosthetic arm, we first check cognitive ability to view it as a cognitive system [2]. Prosthetic arm can be treated as a cognitive system since it can learn, memorize, recognize, and respond to cues. We propose to develop conceptual spaces of prosthetic arm functionalities using geometrical framework of conceptual spaces for the following reasons:

I. To demonstrate the applicability of conceptual spaces to regard prosthetics as an independent cognitive system

II. To demonstrate the utility of conceptual spaces in illustrating the prosthetic system

III. This demonstration might well assist for better vision and insights on prosthetics arm

Gärdenfors suggest that symbolic representation and associations are two significant approaches to describe conceptual spaces [3]. However, concepts are related to each other based on similarities [3, 4]. Hence, geometric structures were introduced to describe conceptual spaces [3]. Geometric spaces describe about the concept using interrelated multidimensional attributes provided each concept is plotted in a geometric space say a multidimensional plane. Concepts tie different properties under different dimension together where each dimension holds different properties of the concepts. Gärdenfors insists that similarities play a major role in cognitive phenomenon [3, 4]. In general, conceptual spaces represent objects and attributes of real world and their relations. It is also possible to represent action and their functional properties by addition force dimension to the conceptual spaces [5]. It is said that force dimension is important to represent quality dimension in learning action and function properties [5, 6]. Concepts are static in nature [3]. For example, apple, car, phone, etc., It is important to add dynamic properties to represent actions. For example, walking, running, throwing, etc., Movements contains sufficient information to perceive underlying force patterns [7]. To identify the structure of action, it is important to identify the similarities between the action involving force patterns. Adding force dimension can help to represent actions [7, 8]. These forces can be physical, emotional or social forms. Geise & Poggio describes action is composed of form and motion pathway. Form pathway consists of sequence of snapshots while motion pathway is the sequence of optic flow. Forces are always experience through interactions [7].

Prosthetics is an artificial system or a device that replaces the human lost or malfunctioning body parts [9]. Prosthetics can include a wide range of devices such as pacemaker, dental implants, limbs, ear implants and so on [10, 11]. Our interest is on prosthetic limbs in particular, upper prosthetic limbs. Upper limbs or hands are crucial parts of human body since it is required for multiple tasks. Losing upper limb could significantly decrease an individual's function in their life. Prosthetics arm could recover certain crucial task for an individual. Prosthetics can be classified into three categories: non-functional prosthetics (aesthetics purpose), mechanically controlled prosthetics and body controlled prosthetics [13]. Among these types body controlled prosthetics are complex to handle as involves many interrelated system such as biological system- the human body, mechanical system- the prosthetic arm and the computerized system- the application software that controls the arm based on the signal from the human body. Body controlled prosthetics arm receives the input electromyography signals (EMG) from via the implanted electrodes. From these raw EMG signals, information is extracted and it is processed further to simulate the action [13].

Human brain records any input from the sensory organs in the form of concepts similar to concepts in conceptual spaces [14]. These concepts can be from by input in the environment. For example, it could a scenery or it be cycling. However, there exist links between each diverse concept in human mind. In that case, the actions performed by human limbs also contribute concepts to conceptual space of the human mind. Inspiring this idea, we propose create a conceptual space of a prosthetic arm. Conceptual space of the prosthetic arm provides better understanding of tasks performed and the relations between the tasks. We adapt the geometrical framework of conceptual spaces to prosthetic arm. We make use of concept lattices in geometric framework of conceptual spaces to achieve cognition in prosthetic arm functionalities. The second section of the paper explains the geometric structure of conceptual spaces. The section 3 of the paper discusses the proposed approach of conceptual space of the prosthetic arm and its and justification of treating prosthetic arm as a cognitive system followed by analysis in section 4. In the later section of the paper, we provide our insights on conceptual space for bidirectional prosthetic arm may possibly develop intelligent prosthetic arm functioning in a way similar to human arm.

**2. Basic terminologies of Conceptual Space:**

As mentioned in previous section, cognitive systems can be represented in three different ways namely symbols, associations and geometrical structures. Geometric framework of the conceptual spaces addresses similarities between the concepts to achieve cognition via geometric structures [3, 4]. Conceptual spaces consist of set of tied concepts under different relation. Each concept is a pair consisting of object and attribute. Objects are real world entities and attributes are the set of properties describing the object. Geometric framework of conceptual spaces holds two prime properties and notion supporting the properties.

*2.1 Properties of conceptual spaces:*

I. *Criterion P: A "natural concept" is a convex region of a conceptual space.* The criterion P says that if an object *O* located between pair of pointe $v_1$ and $v_2$ own some relation with attributes in concept C then all the objects located between the points $v_1$ and $v_2$ also own the attributes possessed by the object *O*.

II. *Criterion C: A concept is represented as a set of convex regions in a number of domains together with information about how the regions in different domains are correlated.* The criterion C says that an Object *O* can be described with attributes from more than one category. This gives rise to Prototype theory. Certain objects are judged to be more representative of an attribute category than others. The most representative member of a category is called prototypical member of that category.

*2.2. Notions of conceptual spaces:*

  I. *Quality dimensions:* Quality dimensions include properties of a real world objects in different domains. For example, temperature, brightness, colour, weight, etc., can be categorized as quality dimensions of an object.
 II. *Domain:* A domain is set of integral and non-separable properties of an object in a dimension. However, it is separable from other dimensions.
III. *Property:* Property forms a region in conceptual spaces.
IV. *Prototype:* Among a group of objects, certain objects are more representative of the category than other. The most representative object of a category is called a prototype.
 V. *Concept:* A concept is a collection of properties falling under different domains.
VI. *Similarities:* Distance between the objects [3] provides the similarity between the objects. Lesser the distance, more the objects are similar to each other.
VII. *Conceptual space:* A conceptual space is the collection of quality dimension that are grouped under different interrelated concepts.

The above subsection explains each notion of conceptual spaces. However, a detailed explanation with illustration can be found [2].

Analogy reasoning involves a general understating of adapting relational information that already exist in one domain memory to another application domain [4]. In this context, similarity is the implied information from source domain to target domain. Three processes appear crucial in the process of analogy reasoning. Firstly, gain access to appropriate source domain information followed by mapping the analogue information from the source domain to target domain. Thirdly, apply of analogue information to produce rules and representations in the target domain. Adapting the properties and notion of geometric framework of conceptual spaces to prosthetic arm functionalities, we propose a conceptual space of prosthetic arm with inspiration from aforementioned analogy reasoning theory. The novelty of the proposed work is adapting the geometrical framework of the conceptual spaces to prosthetic arm to achieve cognition. To our knowledge, very limited literature is available on the applications of conceptual spaces. In this work, we have made use of conceptual lattices instead of voronoi partitioning for achieving similarities in geometric framework of conceptual spaces [15]. Adding to the novelty, we have represented dynamic actions of the prosthetic arm in our concepts using force patterns while in general concepts describe objects. Further, we have used grasp taxonomy attributes to describe the different grasping actions of prosthetic arm functionalities [16]. By developing conceptual space for prosthetic arm, we achieve cognition by exploring similarities of concepts in the conceptual space. This leads to better understanding of prosthetic arm functionalities and their relations. In real time one action is performed in continuous to another can be of ease by adding or removing one or more force patterns instead of bringing prosthetic arm to steady state and starting next action from scratch. In the following subsection, we provide our justification for representation of conceptual spaces via concept lattices and achieving similarity in via concept lattice traversals.

*2.3 Concept lattices for representation of conceptual spaces:*

Philosophical level of study of concepts reveals human knowledge in hierarchical structure [19]. Lattice structures are such hierarchical geometric structures that have been conventionally opted for representing conceptual spaces. Visualising the structures of data by geometric structures is the underlying idea of data analysis [19]. Geometric representation of conceptual spaces can be viewed as collection of concepts in a plane such that subconcept-superconcept-relation corresponds to relation between the concepts in conceptual space. Adam and Raubal [20] suggest that connectionist methods of representing cognitive process adapt mathematically represented nodes and connections. Literature argues lattice structures best represents the conceptual spaces [20]. Considering these, we adopt lattice structures for visualization and representation of conceptual spaces. Gärdenfors claims that concepts are convex regions in multidimensional domains. We regard the concept lattice are convex in 2-dimensional plane with objects and attributes falling under different domain as their dimensions. Another interesting relation between the geometric framework of conceptual spaces and concept lattices is that they hold the similar concepts close to each other while preserving the hierarchies in the concepts [21] [22]. This further justifies the use of concept lattices to represent the conceptual spaces.

*2.4 Similarity in concept lattices:*

Lattices structures exhibits hierarchical ordering of concepts nodes connected with edges [19]. Mathematically, lattice structures exhibits partial ordered relation among them. Each node (concept) except the root and leaf node in the lattice has a super set and sub set. Super sets exist in one level higher than the regarded level while the subset exist one level lower. A concept in a higher level has more objects and less attributes while the concepts at the lower level has less objects and more attributes. This can be viewed as generalization of concepts towards the top and specialization of concepts towards the bottom of the lattice [22]. This approach of traversing concepts can produce most relevant concepts in case of concept of interest is missing or in the case of grouping similar concepts [23]. This implies concepts that are in adjacent levels are more similar than further levels similar to geometric framework of conceptual spaces holding similar concepts with lesser distance between them. The process of generalization reveals the similarity between the concepts while the specialization reveals uniqueness in concepts. Further, the concepts that exists in adjacent levels are similar than the farther level concepts [22] [23]. In this work, we adapt this approach to regard similarity between the concepts.

The concepts and concept lattices root from Formal Concept Analysis. Literature providing in detail explanation about basis of formal concept analysis, concept and concept lattices are available [25-27]. Cognitive concept learning models uses Euclidean metrics to select concept based on concept similarity [28]. It is necessary to establish the cognitive relation between the learnt information. This establishment of relation between the learnt information is required for intelligent behaviour of the cognitive system [29]. Z-numbers are metrics describes cognitive informatics and their reliability. Z- numbers act as a distance measure between the information in a cognitive system [30]. Several possible values of multi-dimensional attributes were considered to perform decision making on uncertain system [31]. Decoding human brain patterns were performed anatomical pattern analysis to categorize visual stimuli activating specific set of regions [32]. Certain measurement models establish frameworks that can measures the behavioural output of representation of mental information [33]. Theories on semantic knowledge says that human perform domain oriented learning rather than individual instance learning in initial years of life [34]. A formal model provides a decomposition of cognitive space into perceptual and conceptual spaces which further explains the human behaviour [35]. Cognitive computational framework generalises the input cues to have better creativity [36]. Geometric representation of correlation explains the similarity between the concepts in terms of correlations. This framework also explains the correlation between attributes of different domains that are held together by the concepts [37]. Learning process has two steps in humans namely, concept learning and categorization of concepts. Concept learning involves storing the processed and reduced dimension information obtained from sensory inputs. Categorization involves identifying the covariance in different domains in order to attain the categories of concepts [38]. From a different perspective, conceptual spaces can be looked a bridge between different forms of information [39].

Following the detailed literature analysis, we propose to model lattice represented conceptual space of prosthetic arm functionalities to actions by adapting geometric framework of conceptual spaces in the next section.

**3. Proposed work:**

Prosthetic arm can be treated as a cognitive system since it can learn, memorize, respond to cues, recognize and distinguish cues. A prosthetic arm can learn action based on the processed input sEMG signal and memorize the set of movements to accomplish the action [9]. Further, prosthetic arm respond by simulating actions for input sEMG signal [13]. It can recognize different input sEMG signal and simulate corresponding action form its input signal. As mentioned earlier, we propose to treat prosthetic arm as a cognitive system to develop the conceptual spaces inspiring analogy reasoning theory [15]. To develop conceptual spaces of a cognitive system, we adapt the formal steps as shown in Figure 1.

The proposed formal method first assesses the cognitive abilities of prosthetic arm. We regard the body controlled prosthetics as prosthetic arm. The considered cognitive functions of prosthetic arm are learning sEMG signals, responding to signals, memorizing the set of movements to accomplish the identified task via sEMG signal and categorization of information. Upon assessment of prosthetic arm possessing the desired cognitive abilities, we adapt the geometrical framework of conceptual spaces to prosthetic arm. Adapting framework implies the deriving the analogy between prosthetic arm component, exercise and actions with regard to Ninapro dataset [16] with the notions of geomteric framework as shown in Table 1. Upon deriving analogy, a definite sample space of prosthetic arm is selected and the notions of geometric framework of conceptual spaces are further adapted. The definite sample space contains the exercises that can be performed with prosthetic arm is chosen. This sample space is decoded in the form data table with objects against attributes and their relations. In this definite sample of prosthetic exercise, we regard the exercises as objects while the components and force patterns performed by the components as attributes of the decoded data table which is the input for concept generation algorithm shown in Figure 2. This concept generation algorithm generates the list of concepts that forms the basis for the conceptual spaces of the prosthetic arm. Upon generation of concepts, the lattice representation of conceptual space is built using the algorithm shown in Figure 3. The lattice representation of conceptual spaces is similarity explored further for achieving cognition [3] [4]. Figure 2 and 3 represents the GRAIL algorithm implanted in ConExp [40].

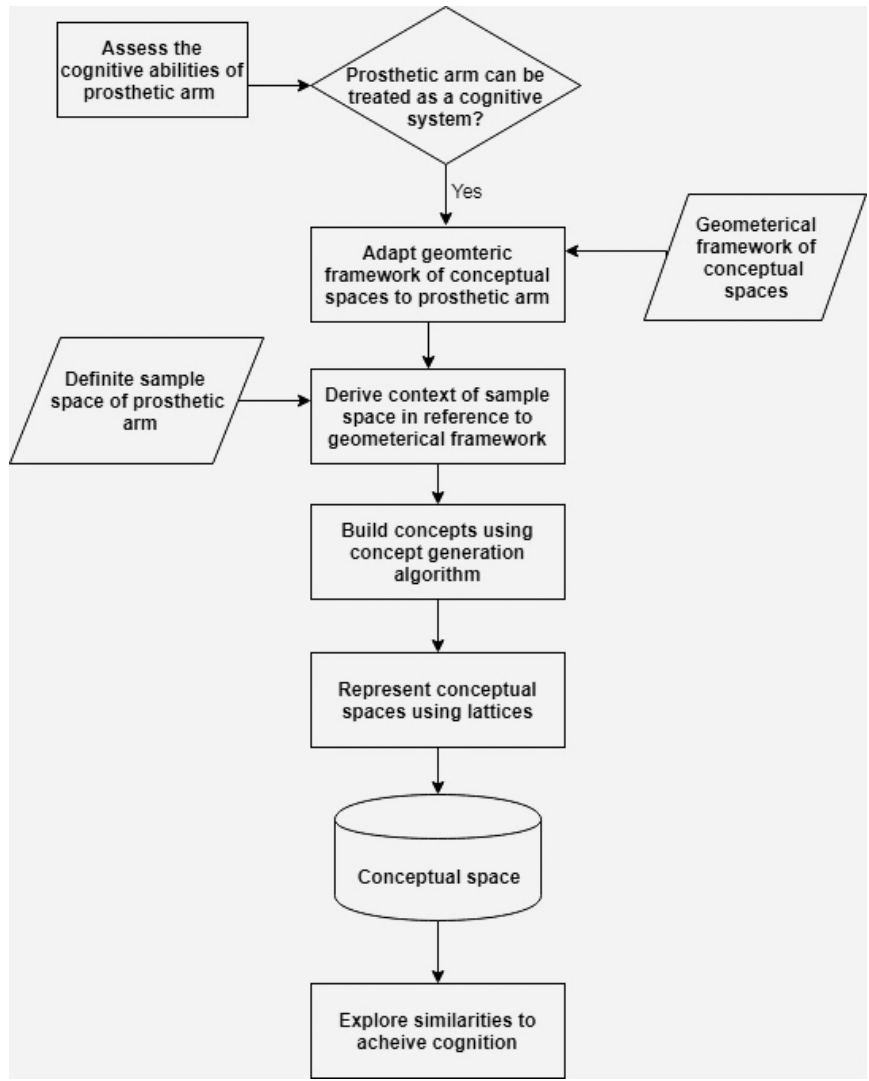

Figure 1: Formal method for developing conceptual spaces of Prosthetic arm

Using the proposed method, we develop the lattice represented conceptual space of the prosthetic arm functionalities. In this lattice represented conceptual spaces we exploit the traversals of lattices to obtain similar concepts. We have made use to concept lattices with geometric framework of conceptual spaces to overcome individual limitations of concept theories such as context dependency and vagueness respectively.

Table 1: Conceptual spaces of the Prosthetic arm

| S. No | Conceptual spaces | Conceptual spaces of the prosthetic arm | | | |
|---|---|---|---|---|---|
| 1 | Quality dimension | Fingers, Wrist, Forces, Grasp | | | |
| 2 | Domain | Fingers | | Forces | Grasp |
| | | Index, Middle, Ring, Little, Thumb | Wrist | Abduction, Flexion, Extension, Adduction, Up, Flexed, Over, Opposing, base, Point, Close, Rotate, Deviation | Power, Intermediate, Precision, Pad, Palm, Side, VF1, VF2, VF3 |
| 3 | Property | Each exercise has a property, for example wrist exercises has the property of angular rotation. | | | |
| 4 | Prototype | Most representative object of an exercise category. For example, Under grasp exercise, holding a ball is more representative when compared to grasping exercise. | | | |
| 5 | Concept | 52 exercise of Ninapro dataset | | | |
| 6 | Similarities | If action are said to be similar, the Euclidean distance between the concepts should be minimal. For example, 'waving' and 'shaking' can be similar that 'waving' and 'throwing'. | | | |
| 7 | Conceptual spaces | The set of all interrelated exercise (concepts) forms the conceptual space of the prosthetic arm. | | | |

```
Input: Decoded sample space
Output: Set of concepts

C=(G,G')
if(M≠G')
    CalcPredConcepts((G,G'), ∅)
end if
CalPredConcepts((A',A"), Prohibited)
    Prohibited=Prohibited ∪ A"
    For each m ∈ M\Prohibited
        NewIntent= M∩{∩g^I|g∈A' & m∈g^I}
        NewExtent={g|g∈A' & m∈g^I}
        Outer={∪g^I|g∈A' & m∉g^I}
        if(NewIntent∩Prohibited)\A"= ∅
            C=C∪ (NewIntent. NewExtent)
            CalPredConcepts((NewIntent,NewExtent), Prohibited)
            Prohibited= Prohibited ∪ (M\Outer)
        end if
```

Figure 2: Concept Generation Algorithm of the proposed model

```
Input: Set of concepts
Output: Lattice represented conceptual space
C=(G,G')
if( G,G') ≠ (M', M)
    C=C ∪ (M',M)
    Findpredecessors((G,G'), ∅, ∅)
end if
FindPredecessors((A',A"),Prohibited)
    Desc=(∪{g^I, g∈A'})\A")
    if Desc=∅
        Connect((A',A"),(M',M))
    else
        WorkSet=Desc
    end if
    for each m ∈ Workset
        Intent= M ∩(∩{g^I|g∈ A' & m∈g^I})
        Extent= {g|g∈A' & m∈g^I}
        Outer= ∪{g^I| g∈A' & m ∉ g^I}
        if (Intent ∩ Outer)\A''=∅
            WorkSet={WorkSet\Intent}∩Outer
            if (Intent ∩ Prohibited)\A''=∅
                if Intent=M
                    Connect((A',A"),(M',M))
                else
                    C=C∪(Extent, Intent)
                end if
            end if
        end if
            FindPredecessors((Extent, Intent), Prohibited)
            Prohibited=Prohibited∪{m}
```

Figure 3: Algorithm for building Concept Lattice

Concept lattices are mental construct form of conceptual spaces [20]. The set of attributes and their values representing the objects is composed under an indivisible entity concept. These concepts ordered in semantic network obeying inheritance forms concept lattices [20]. Hence, concept lattices are best suitable for representing conceptual spaces. In the next section of the paper, we develop the conceptual spaces of prosthetic arm functionalities using the proposed model. In the next section of the paper, we first develop the conceptual spaces using force patterns. Secondly, certain exercises of definite data sample are grasp exercises. The grasp exercises of the prosthetic arm requires more than the force patterns. We introduce the grasp taxonomy and its attributes for grasp functionality of prosthetic arm in the analysis section of the paper. As mentioned, we have provided a detailed illustrative analysis with Ninapro dataset [16] by adapting the proposed method.

**4. Illustrative Analysis of conceptual spaces of Prosthetic arm:**

This section performs a detailed analysis of proposed model on NinaPro dataset to create conceptual space of the defined definite sample. This dataset is considered for analysis since it has the highest number of action and subject [16] and can be mimicked by prosthetic arm with ease.

NinaPro dataset provides surface electromyography signals of around 50 hand actions performed by 67 intact subjects and 11 amputated subjects. Surface EMG is collected from the subject's surface of skin by collecting the force pattern of interest. It is worth noting that sEMG is related to forces rather than positions [16]. The actions of Ninapro dataset were performed by intact subject were also performed by amputated subjects.

Ninapro dataset has four exercises namely finger movements, hand postures, wrist movements and grasping and functional movements. Each exercise is a hand gesture with movements and can be represented with force attribute involved in the action. The NinaPro dataset consist four exercise sections namely 12 basic hand movements, 8 isometric and isotonic hand configuration, 9 basic wrist movements and 23 basic grasping and functional movements.

By creating the conceptual spaces for the prosthetic arm functionalities, we can explore the similarities and difference between each action of the prosthetic arm. This in turn, can help us to achieve moving from one action to another action with ease rather performing actions from base in real time since most actions are continuous in real time. By performing reasoning in prosthetic arm functionalities we could have better understanding on relation between functionalities. Further, in real time performing one action in continuous to another can be of ease by adding or removing one or more force patterns. This can possibly avoid bringing prosthetic arm to steady state and starting next action from scratch whenever possible. Also, once could learn an action just adding one or set of force patterns.

As mentioned in previous sections, we represent conceptual space with conceptual lattices since concept lattices and geometric framework of conceptual spaces holds similar concepts to closer. In order to build a concept lattice, we first create the set of concepts followed by ordering concepts in concept lattice.

Upon selecting the definite sample of prosthetic arm, we decode the dataset into data table with exercises as objects, components as well as force patterns performed by the components as attribute and their relations. Upon decoding, concepts are formed using algorithm in Figure 2. The concept lattice is created using the algorithm in Figure 3. Upon creating concept lattice, similarity is achieved as explained in section 2.4. Detailed explanation of each step is provided below.

**4.1 Concept Formation:**

The definite sample is decoded as the data table with experiments performed under different exercise as per NinaPro dataset as object while the attribute are the actions performed by the components of the prosthetic arm. The components of a prosthetic arm of our scope are Wrist, Index finger, Middle finger, Ring finger, Little finger and Thumb. The actions performed by these components include Flexion, Extension, Adduction, Abduction, Open up, Flexed over, Oppose base, Point, Close, Rotate and Deviate as shown in Figure 4. The components and the actions together are regarded as the attributes of the data table. We have not included the grasp exercise of the Ninapro dataset in this section since grasp cannot be just performed with force pattern. It requires force in addition to grasp attributes of grasp taxonomy [41-43]. We have explained the same in the later part of the current section. The data table is formed as mentioned is shown in Figure 5 of the paper.

**4.1.1 Concept Learning:**

Concepts are learned from the aforementioned data table using concept generation algorithms. There are numerous algorithms in literature for concept learning. We choose to use the concept learning algorithm shown in Figure 2. This algorithm learns the list of all possible concepts without redundancy from the data shown in Figure 5. These concepts record the meaningful grouping and categorization of exercises with components and actions. This list of concept is the input for the lattice creation algorithm shown in Figure 4 to obtain the lattice represented conceptual space of the definite sample.

Table 2: List of objects and attributes of the conceptual space of Prosthetic arm

| Object list: | Attribute list: | |
|---|---|---|
| | Components of prosthetic arm: | Actions: |
| 52 Exercises of NinaPro dataset | Wrist, Index finger, Middle finger, Ring finger, Little finger and Thumb | Flexion, Extension, Adduction, Abduction, Open up, Flexed over, Oppose base, Point, Close, Rotate and Deviate |

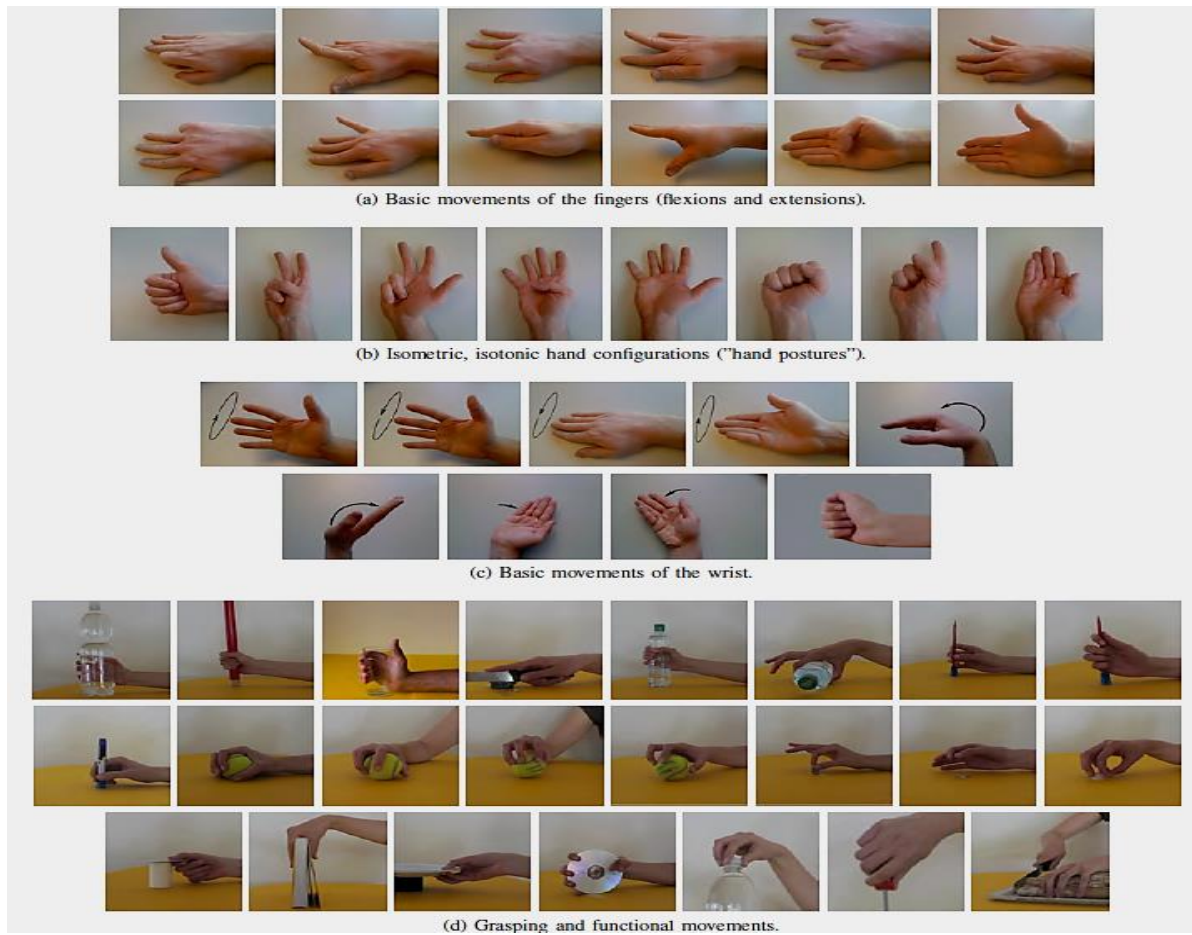

Figure 4: Actions performed in NinaPro dataset [16]

The lattice representation of conceptual space is shown in Figure 6. In this figure each circle is a concept connecting the objects in non-shaded rectangular boxes (exercises) and components as well as force patterns in shaded rectangular boxes (attributes) of the conceptual space. It can be observed from Figure 6 that each exercise (objects) is represented by force pattern as well as components performing the force pattern (attributes) in way similar to concepts in human memory [14]. Further, the arrangement of concepts in Figure 6 shows that concepts within an exercise were held closer than the concepts between the exercises. On taking a closer look, one can observe that concepts belongs belonging to exercise 1 of Ninapro dataset are present at the centre of the conceptual space while exercise 2 and 3 are present towards the right and the left side of the conceptual space. This is because the Exercise 1 of Ninapro dataset (Basic hand movements) are required for both Exercise 2 (Isometric and Isotonic hand configurations) as well as Exercise 3 (Basic Wrist movements). This serves well for the argument that similar concepts are held together in lattice represented conceptual space in a way similar to geometric framework of conceptual space. The concepts that are adjacent are more similar but differ by one or few set of attributes. This leads us to understanding of generalization and differentiation of concept based on set of attributes of interest.

The functions listed exercise A, B and C are explicit and can be reproduced by manipulating joints and force patterns. Exercise D certainly lacks clarity [5] with just force patterns. The exercise D of Ninapro dataset are grasping or holding requiring efforts different that just force patterns. For example, holding a ball is different from holding a knife. They do not differ just by force, there are several other parameters involved [17]. Hence, we introduce grasp attributes for grasp concepts of conceptual spaces of prosthetic arm from grasp taxonomy [41].

**4.1.1.1 Grasp classification:**

Grasp is a hand gesture in which one or more fingers securely hold the object. Figure 7 shows the different grasp types and subtypes with reference to grasp taxonomy. Grasp can be classified based on:

*a. Grasp intention:*

Based on the underlying intention, grasp may be classified into power grasp, intermediate grasp and precision grasp. Power grasp holds a rigid relation between object and hand. All movements of the objects have to be from the arm. While precision grasp does not require arm movements and thus movements has to be internal. When an action needs both precision and power equally, it can be categorized under intermediate grasp.

|          | Index Finger | Middle Finger | Ring Finger | Little Finger | Thumb | Abduction | Flexion | Extension | Adduction | Up | Flexed over | Opposing base | Point | Close | Wrist | Rotate | Deviation |
|---|---|---|---|---|---|---|---|---|---|---|---|---|---|---|---|---|---|
| Ex 1 Act 1- | 1 | 0 | 0 | 0 | 0 | 0 | 1 | 1 | 0 | 0 | 0 | 0 | 0 | 0 | 0 | 0 | 0 |
| Ex 1 Act 3- | 0 | 1 | 0 | 0 | 0 | 0 | 1 | 1 | 0 | 0 | 0 | 0 | 0 | 0 | 0 | 0 | 0 |
| Ex 1 Act 5- | 0 | 0 | 1 | 0 | 0 | 0 | 1 | 1 | 0 | 0 | 0 | 0 | 0 | 0 | 0 | 0 | 0 |
| Ex 1 Act 7- | 0 | 0 | 0 | 1 | 0 | 0 | 1 | 1 | 0 | 0 | 0 | 0 | 0 | 0 | 0 | 0 | 0 |
| Ex 1 Act 9- | 0 | 0 | 0 | 0 | 1 | 1 | 0 | 0 | 1 | 0 | 0 | 0 | 0 | 0 | 0 | 0 | 0 |
| Ex 1 Act 11 | 0 | 0 | 0 | 0 | 1 | 0 | 1 | 1 | 0 | 0 | 0 | 0 | 0 | 0 | 0 | 0 | 0 |
| Ex 2 Act 1 | 0 | 0 | 0 | 0 | 1 | 0 | 0 | 0 | 0 | 1 | 0 | 0 | 0 | 0 | 0 | 0 | 0 |
| Ex 2 Act 2 | 0 | 1 | 0 | 1 | 1 | 0 | 1 | 0 | 0 | 0 | 1 | 0 | 0 | 0 | 0 | 0 | 0 |
| Ex 2 Act 3 | 0 | 0 | 1 | 1 | 0 | 0 | 1 | 0 | 0 | 0 | 0 | 0 | 0 | 0 | 0 | 0 | 0 |
| Ex 2 Act 4 | 0 | 0 | 0 | 1 | 1 | 0 | 0 | 0 | 0 | 0 | 0 | 1 | 0 | 0 | 0 | 0 | 0 |
| Ex 2 Act 5 | 1 | 1 | 1 | 1 | 1 | 0 | 1 | 0 | 0 | 0 | 0 | 0 | 0 | 0 | 0 | 0 | 0 |
| Ex 2 Act 6 | 1 | 1 | 1 | 1 | 1 | 1 | 0 | 0 | 0 | 0 | 0 | 0 | 0 | 0 | 0 | 0 | 0 |
| Ex 2 Act 7 | 1 | 0 | 0 | 0 | 0 | 0 | 0 | 0 | 0 | 0 | 0 | 0 | 1 | 0 | 0 | 0 | 0 |
| Ex 2 Act 8 | 1 | 1 | 1 | 1 | 1 | 0 | 1 | 0 | 0 | 0 | 0 | 0 | 0 | 1 | 0 | 0 | 0 |
| Ex 3 Act 1- | 0 | 1 | 0 | 0 | 0 | 0 | 0 | 0 | 0 | 0 | 0 | 0 | 0 | 0 | 1 | 1 | 0 |
| Ex 3 Act 3- | 0 | 0 | 0 | 1 | 0 | 0 | 0 | 0 | 0 | 0 | 0 | 0 | 0 | 0 | 1 | 1 | 0 |
| Ex 3 Act 5- | 0 | 0 | 0 | 0 | 0 | 0 | 1 | 1 | 0 | 0 | 0 | 0 | 0 | 0 | 1 | 0 | 0 |
| Ex 3 Act 7- | 0 | 0 | 0 | 0 | 0 | 0 | 0 | 0 | 0 | 0 | 0 | 0 | 0 | 0 | 1 | 0 | 1 |
| Ex 3 Act 9 | 1 | 1 | 1 | 1 | 1 | 0 | 0 | 1 | 0 | 0 | 0 | 0 | 0 | 1 | 1 | 0 | 0 |

Figure 5: Data table of NinaPro dataset of exercise A,B and C.

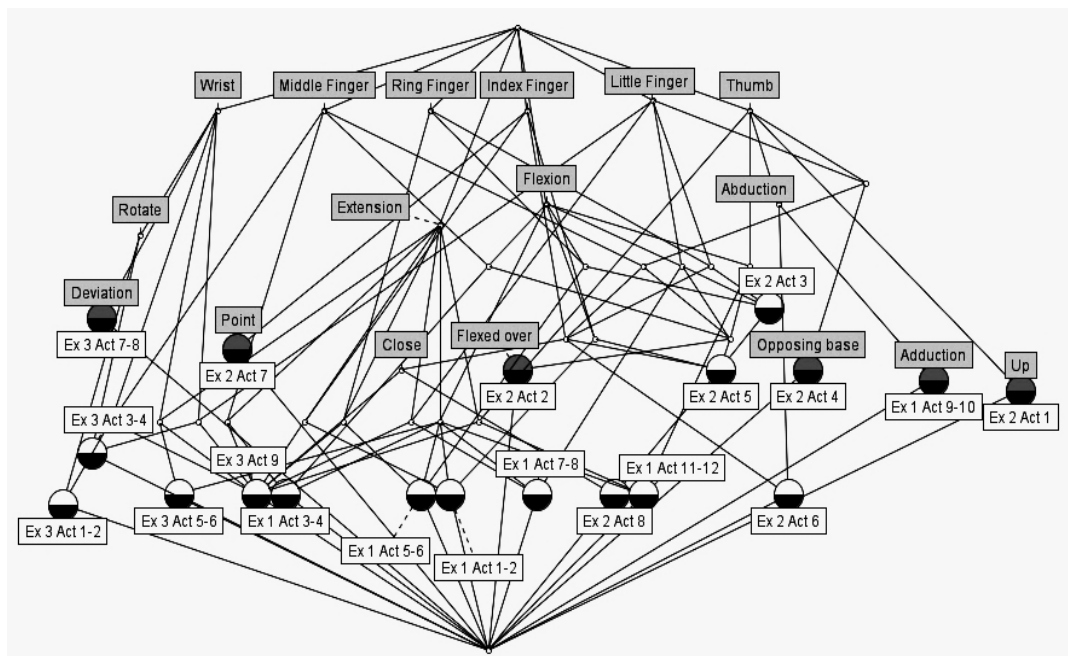

Figure 6: Conceptual space of Ninapro dataset of Exercise A, B and C

*b. Opposition types:*

Based on the opposition type, the grasp can be classified into pad opposition, palm opposition and side opposition as shown in Figure 8. In pad opposition grasping, the surface of the hand generally lies parallel to palm. In palm opposition grasping, the surface of the hand generally lies perpendicular to palm. Side opposition grasp has surface of the hand transverse to palm.

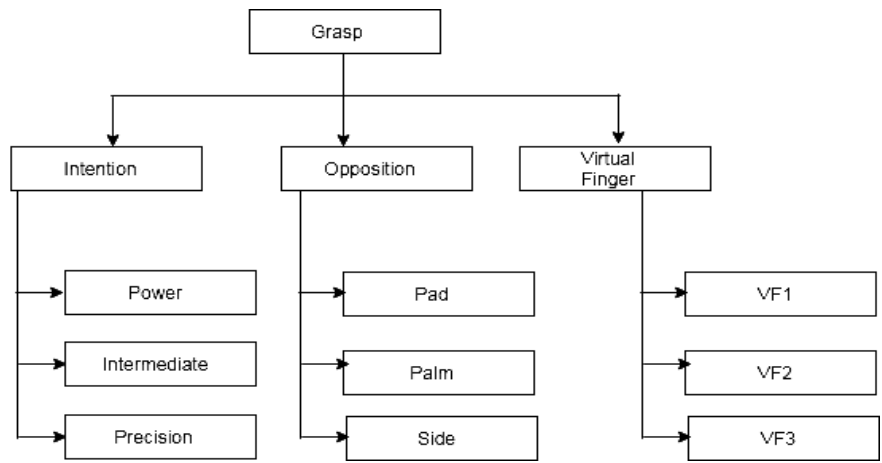

Figure 7: Different types of Grasps

*c. Virtual Finger:*

In a grasp function, several components of hand work together as a virtual finger. A unit belong to virtual finger if the unit apply force in the same direction as its coordinating unit. The opposition of palm against hand, opposition of fingers against hand and opposition of finger against force can be categorized as VF1, VF2 and VF3.

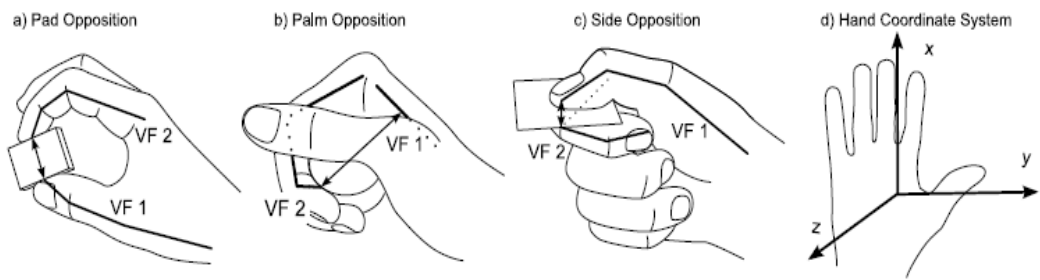

Figure 8: Different types of grasp as per Grasp Taxonomy [43]

We incorporate attributes such as intention of grasp, opposition type of grasp and virtual finger to identify different grasping exercise of Ninapro dataset. Figure 5 and 6 deals with only three exercises of Ninapro dataset. The fourth grasp exercise of the dataset is explained with the aforementioned grasp attributes as per grasp taxonomy instead of just force attribute.

On applying the grasp attributes to the exercise four of Ninapro dataset, we obtain a data table of grasp exercises as shown in Figure 9 for grasp exercise (Exercise 4 of Ninapro dataset) to obtain the list of concepts using the algorithm shown in Figure 3. From the obtained set of concepts, we create the lattice represented conceptual space of grasp functionalities of prosthetic arm using the algorithm shown in Figure 4.

The lattice represented conceptual space of prosthetic arm for grasp functionalities is shown in Figure 10. In Figure 10, concepts are represented in circles which connect the grasp exercises of Ninapro dataset with grasp attributes as well as components of the prosthetic arm. Similar exercises are group together as well as held closer while the exercises that are not similar are held farther in reference to grasp taxonomy and Ninapro dataset [43, 16].

Figure 6 and Figure 10 show the lattice represented conceptual spaces. We have made use of ConExp [40] for concept generation and lattice creation. ConExp adapts concept generation and lattice creation algorithm shown in Figure 3 and 4 respectively. The concepts are arranged in the hierarchical order in which each concept of interest has super concept and sub concept except root and leaf concepts. It is trivial that concepts that are similar are grouped together under a same node.

To explain how similar concepts are held closer in lattice represented conceptual space in a way similar to geometric framework of conceptual space, we have provided an example for each case. We have considered four cases of holding similar concepts closer in this paper namely: Superconcept similar to concept of interest, Subconcept similar to concept of interest, Sibling concept similar to concept of interest, similar concepts.

The super concept and subconcept of a concept is much similar to a concept of interest. Super concept of a concept of interest is quite generalized in terms of attributes while it is specialized in terms of objects. In other words, the subconcept of a concept of interest is quite specialized in terms of attributes while it is generalized in terms of objects. Sibling concepts are next similar concepts with an identified variation. The concepts that are near in a lattice are much similar than the concept that are far in a way similar to euclidean metric of concept discretization. In the following, we pick an example for each case to support our analysis of concept similarity.

*Case 1- Super concepts are much similar to concept of interest:*

The concept Ex 3 Act 6 is the super set of the concept Ex 3 Act 9. Ex 3 Act 6 is wrist extension while Ex 3 Act 9 is Wrist extension with closed hand. Wrist extension is similar to wrist extension with closed hand than Wrist rotation.

*Case 2- Subconcepts are much similar to concept of interest:*

The concept Ex 2 Act 6 is the subconcept of concept Ex 2 Act 8. It can be noted that flexion of all fingers (Ex 2 Act 6) is much similar to fingers closed together (Ex 2 Act 6) than flexion of little finger and ring finger (Ex 2 Act 3).

|        | Power | Intermediate | Precision | Pad | Palm | Side | VF1 | VF2 | VF3 | Abduction | Adduction |
|--------|-------|--------------|-----------|-----|------|------|-----|-----|-----|-----------|-----------|
| Ex4 Act | 1 | 0 | 0 | 0 | 1 | 0 | 1 | 1 | 0 | 1 | 0 |
| Ex4 Act | 1 | 0 | 0 | 0 | 1 | 0 | 1 | 1 | 0 | 0 | 1 |
| Ex4 Act | 1 | 0 | 0 | 0 | 1 | 0 | 1 | 1 | 1 | 0 | 1 |
| Ex4 Act | 1 | 0 | 0 | 0 | 1 | 0 | 1 | 1 | 0 | 1 | 0 |
| Ex4 Act | 1 | 0 | 0 | 1 | 0 | 0 | 1 | 1 | 0 | 1 | 0 |
| Ex4 Act | 0 | 0 | 1 | 1 | 0 | 0 | 1 | 1 | 0 | 1 | 0 |
| Ex4 Act | 0 | 1 | 0 | 0 | 0 | 1 | 1 | 1 | 0 | 0 | 1 |
| Ex4 Act | 0 | 0 | 1 | 0 | 0 | 1 | 1 | 1 | 0 | 1 | 0 |
| Ex4 Act | 1 | 0 | 1 | 1 | 1 | 0 | 1 | 1 | 0 | 1 | 0 |
| Ex4 Act | 0 | 0 | 1 | 1 | 0 | 0 | 1 | 1 | 0 | 1 | 0 |
| Ex4 Act | 0 | 0 | 1 | 1 | 0 | 0 | 1 | 1 | 0 | 1 | 0 |
| Ex4 Act | 0 | 0 | 1 | 1 | 0 | 0 | 1 | 1 | 0 | 1 | 0 |
| Ex4 Act | 0 | 1 | 0 | 0 | 0 | 1 | 1 | 1 | 0 | 0 | 1 |
| Ex4 Act | 0 | 0 | 1 | 1 | 0 | 0 | 1 | 1 | 0 | 0 | 1 |
| Ex4 Act | 1 | 0 | 0 | 0 | 1 | 0 | 1 | 1 | 0 | 1 | 0 |
| Ex4 Act | 0 | 0 | 1 | 1 | 0 | 0 | 1 | 1 | 0 | 1 | 0 |
| Ex4 Act | 0 | 1 | 0 | 0 | 0 | 1 | 1 | 1 | 0 | 1 | 0 |
| Ex4 Act | 1 | 0 | 0 | 0 | 1 | 0 | 1 | 1 | 1 | 0 | 1 |

Figure 9: Grasp data table of the prosthetic arm

*Case 3- Sibling concept similar with an identified variation:*

The concepts Ex 1 Act 1-2, Ex 1 Act 3-4, Ex 1 Act 5-6, Ex 1 Act 7-8 and Ex 1 Act 11-12 are sibling concepts which performs flexion and extension of fingers. However, Ex 1 Act 1-2 is flexion and extension of index finger while Ex 1 Act 3-4, Ex 1 Act 5-6, Ex 1 Act 7-8 and Ex 1 Act 11-12 performs on middle, ring, little finger and thumb respectively. The aforementioned exercises perform flexion and extension but on different fingers.

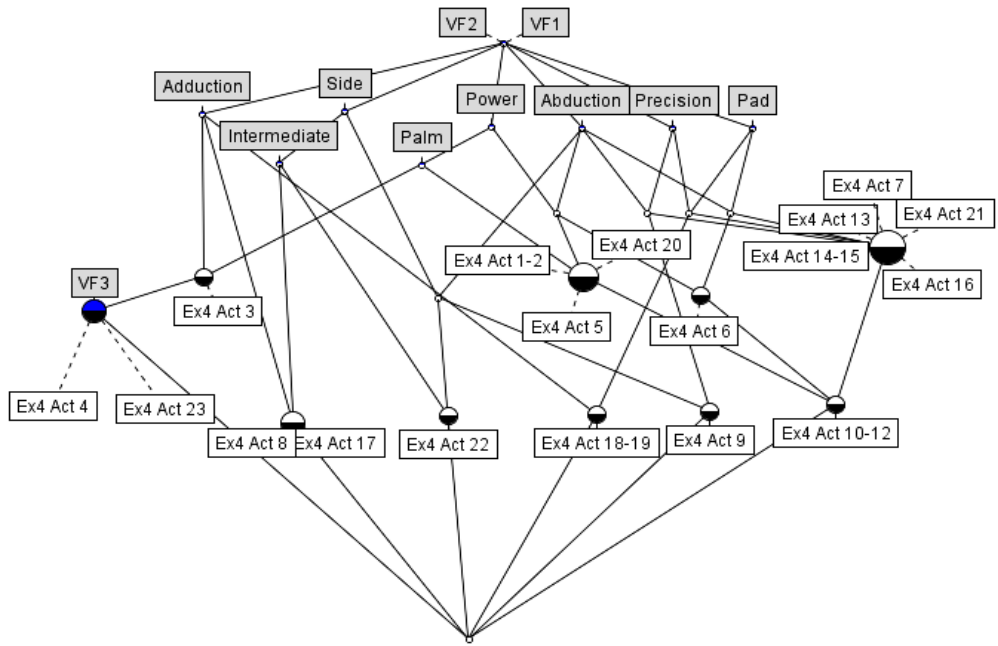

Figure 10: Grasp conceptual space of the prosthetic arm

*Case 4- Similar concepts:*

The concept Ex 4 Act 4 is similar to concept Ex 4 Act 23. It can be noted that concept Ex 4 Act 4 is grasp with index finger extension while Ex 4 Act 23 is grasping a knife with index finger extension.

The analysis of proposed conceptual spaces is performed with four different cases namely super concept is similar to concept of interest, sub concept similar to concept of interest, sibling concept similar with identified variation and similar concepts. Further, this also shows the level of similarity between the level of concept lattice. Case 1 and 2 explains the similarity existing in super concepts and subconcepts. Similarly, sibling concepts exist in same level with an identified variation. Case 3 explains the sibling concepts similarity. There are certain concepts that are similar and can be grouped under single concepts. Case 4 explains such similar concepts. The above four cases explains achieving cognition via reasoning based on similarity and differences [15].

In this paper, we have developed the conceptual space for prosthetic arm functionalities by adapting the geometric framework of conceptual space and concept lattice of Formal Concept Analysis. The unified concept theory integrates the theories of concepts [44]. FCA and geometric framework of conceptual spaces are one among theories of unified concept theory. Individual concept theories have one or more limitations. Geometric conceptual spaces have a limitation of vagueness while the FCA has a limitation of context dependency. In this paper, we have integrated FCA with geometric framework of conceptual space to overcome the limitations of one another while attaining cognition via exploring similarities and difference in the obtained conceptual space. Further, we have applied this integrated this approach to application of prosthetic arm functionalities.

## 5. Discussions:

Prosthetics regarded in this paper is body controlled prosthetics dealing with simulation of sEMG signals by electrodes of the prosthetic arm in a way similar to human arm. The sEMG signals are collected from nerves endings and converted by processing signals to actions [12]. These signals are raw data of force pattern and hence we have used force attributes to describe the exercises. It can be noted that there exist a feedback from human arm to human brain based about the performed force of interest and the force is adjusted accordingly. This kind of feedback is not available in normal prosthetic arms. However, a recent research [45] has designed a bidirectional prosthetic arm. This bidirectional prosthetic arm utilizes the fast growing the neural interfaces to provide feedback to the peripheral nervous system. This research has demonstrated that the participants were able to perceive different object, perform different action under blindfold condition. It can be noted that literature suggests that bidirectional communication is much required [45]. If conceptual spaces are adapted to bidirectional hand prosthetics system, it may well demonstrate cognitive phenomena with communication or feedback. By adapting the conceptual spaces and FCA based bidirectional associative memory, we may well attain cognitive abilities such as learning, memorising, recall and reasoning in bidirectional prosthetics system [23]. However, bidirectional prosthesis is growing research discipline and very few literature addresses bidirectional prosthesis. We conclude our contribution in the next section having provided our insights of intelligent prosthetic arm that can provide feedback in addition to achieving cognition.

## 6. Conclusions:

In this paper, we have adapted geometrical framework of conceptual spaces to achieve cognition in prosthetic arm functionalities via reasoning. We regard reasoning as identified similarities and differences in conceptual space lattice. During this reasoning, we have introduced concept space lattices rather than original voronoi partitioning. The intention behind such treatment is to show the power of conceptual spaces, cognitive ability of the prosthetic system and with a vision for betterment in the field of prosthetics. Also, we have provided our insights for conceptual spaces for bidirectional prosthetic arm for our future work.

**Acknowledgement:** Authors sincerely acknowledge the support from Department of Science and Technology, Government of India under the Grant SR/CSRI/118/2014.

**Compliance with Ethical Standards:** This article does not contain any studies with human participants or animals performed by any of the authors.